\documentclass{article}

\usepackage{microtype}
\usepackage{graphicx}
\usepackage{subfigure}
\usepackage{booktabs} % for professional tables
\usepackage{multirow}
\usepackage{diagbox}
\usepackage{float}
\usepackage{placeins} % for \FloatBarrier

\usepackage{hyperref}

\usepackage[accepted]{icml2025}

\usepackage{amsmath}
\usepackage{amssymb}
\usepackage{mathtools}
\usepackage{amsthm}

\usepackage[capitalize,noabbrev]{cleveref}

\theoremstyle{plain}

\theoremstyle{definition}

\theoremstyle{remark}

\usepackage[textsize=tiny]{todonotes}

\icmltitlerunning{Strategic Fusion Optimizes Transformer Compression}

\begin{document}

\twocolumn[
\icmltitle{Strategic Fusion Optimizes Transformer Compression}

% It is OKAY to include author information, even for blind
% submissions: the style file will automatically remove it for you
% unless you've provided the [accepted] option to the icml2025
% package.

% List of affiliations: The first argument should be a (short)
% identifier you will use later to specify author affiliations
% Academic affiliations should list Department, University, City, Region, Country
% Industry affiliations should list Company, City, Region, Country

% You can specify symbols, otherwise they are numbered in order.
% Ideally, you should not use this facility. Affiliations will be numbered
% in order of appearance and this is the preferred way.
% \icmlsetsymbol{equal}{*}

\begin{icmlauthorlist}
\icmlauthor{Md Shoaibur Rahman}{} \\
\texttt{shoaibeee@gmail.com} \\
% \icmlauthor{}{sch}
\end{icmlauthorlist}

% \icmlaffiliation{yyy}{Department of XXX, University of YYY, Location, Country}
% \icmlaffiliation{comp}{Undefined}
% \icmlcorrespondingauthor{Shoaib}{shoaibeee@gmail.com}

% You may provide any keywords that you
% find helpful for describing your paper; these are used to populate
% the "keywords" metadata in the PDF but will not be shown in the document
\icmlkeywords{Transformer Models, Model Compression, Layer Pruning, Knowledge Distillation, Strategic Fusion, Performance Optimization}

\vskip 0.3in
]

% \printAffiliationsAndNotice{}  % leave blank if no need to mention equal contribution
% \printAffiliationsAndNotice{\icmlEqualContribution} % otherwise use the standard text.

\begin{abstract}
This study investigates transformer model compression by systematically pruning its layers. We evaluated 14 pruning strategies across nine diverse datasets, including 12 strategies based on different signals obtained from layer activations, mutual information, gradients, weights, and attention. To address the limitations of single-signal strategies, we introduced two fusion strategies, linear regression and random forest, which combine individual strategies (i.e., strategic fusion), for more informed pruning decisions. Additionally, we applied knowledge distillation to mitigate any accuracy loss during layer pruning. Our results reveal that random forest strategic fusion outperforms individual strategies in seven out of nine datasets and achieves near-optimal performance in the other two. The distilled random forest surpasses the original accuracy in six datasets and mitigates accuracy drops in the remaining three. Knowledge distillation also improves the accuracy-to-size ratio by an average factor of 18.84 across all datasets. Supported by mathematical foundations and biological analogies, our findings suggest that strategically combining multiple signals can lead to efficient, high-performing transformer models for resource-constrained applications.

\end{abstract}

\section{Introduction}
\label{submission}
Large pre-trained transformer models \cite{vaswani2023attentionneed} have transformed natural language processing by achieving state-of-the-art performance across a wide range of tasks, from sentiment analysis to text classification \cite{rahman2024llmvariants}. However, their significant computational and memory requirements present a challenge for deployment in resource-constrained environments, such as edge devices or real-time applications \cite{strubell2019energypolicyconsiderationsdeep}. Model compression has emerged as a crucial area of research to address these limitations, enabling efficient use of these powerful models without compromising their performance.

Past studies explored various approaches for compressing large transformer models, including pruning unimportant weights \cite{han2015learningweightsconnectionsefficient}, quantizing parameters \cite{gong2014compressingdeepconvolutionalnetworks}, and knowledge distillation \cite{hinton2015distillingknowledgeneuralnetwork}. Recent work also extends beyond individual parameters to prune entire layers considered less critical, using strategies such as activation-based \cite{tushar2024activationbasedpruning}, gradient-based \cite{molchanov2017pruningconvolutionalneuralnetworks, yang2023gradientbasedintraattentionpruningpretrained}, mutual information-based \cite{isik2022informationtheoreticjustificationmodelpruning}, weight-based \cite{frankle2019lotterytickethypothesisfinding}, or attention-based \cite{michel2019sixteenheadsreallybetter} pruning. Although these techniques effectively reduce model size and computational costs, they often rely on single metrics to determine layer redundancy.

However, focusing solely on a single signal, which often requires predefined pruning rules, may not fully capture the nuanced contributions of a layer to downstream tasks \cite{hooker2021compresseddeepneuralnetworks, zhang2022platonpruninglargetransformer}. Therefore, single signal-based pruning strategies often lead to drastic drops in accuracy. Crucially, existing works seldom explore how to combine multiple pruning signals or thoroughly examine how to sequence layer pruning decisions without a predefined rule. This gap underlines the need for a framework that accounts for multiple pruning signals and systematically assesses the importance of each layer to optimize performance trade-offs.

In this work, we examined 12 individual pruning strategies using signals from layer activations, mutual information, gradients, weights, and attention, and propose two fusion strategies to integrate these signals. We provided the mathematical and biological intuition behind the choice of each strategy, which demonstrates theoretical and practical perspectives on their selection. We tested our strategies using the BERT model \cite{devlin2019bertpretrainingdeepbidirectional}. We evaluated 14 pruning strategies across nine datasets, focusing on text classification and sentiment analysis. Each individual strategy computes a layer-specific metric or signal, identifies layers to prune based on a predefined rule (e.g., based on the min or max value of the metric), and fine-tunes the compressed model. The fusion strategies, based on linear regression and random forest, integrate multiple pruning signals to automatically identify optimal layer pruning schedules without a predefined rule. Finally, we incorporated knowledge distillation-based training, where the compressed model was trained using the original model as a teacher to recover any accuracy lost during pruning.

Our experiments reveal that integrating multiple signals through strategic fusion consistently outperforms single-metric approaches in both accuracy improvement and model size reduction. In particular, random forest-based fusion strategy achieves the best performance in seven out of nine datasets, while ranking second and third best for the remaining two datasets. Furthermore, knowledge distillation exceeds the original accuracy for six datasets and mitigates the accuracy drops in three other datasets. The accuracy-to-size ratio after distillation increases by an average factor of 18.84 across all datasets. Our results also highlight that which layers are pruned and in what sequence matters greatly: edge layers often carry critical information, and high-performing strategies automatically learn not to prune them early. Taken together, our findings demonstrate that the fusion of individual strategies into a data-driven framework can lead to an effective and efficient compressed transformer model.

\section{Methodology}

\subsection{Datasets}
We employed nine diverse text classification datasets of varying domains (e.g., user reviews, scientific abstracts, news articles) and number of labels (2 to 20 classes): newsgroup, dbpedia\_14 (dbpedia), arxiv-classification (arxiv), patent-classification (patent), yahoo\_answers\_topics (yahoo), yelp\_review\_full (yelp), ag\_news (agnews), imdb, and amazon\_polarity (amazon). All datasets except newsgroup are available via Hugging Face, whereas newsgroup is accessible through scikit-learn. All input sequences were tokenized using the BERT tokenizer and padded or truncated to a maximum length of 32 tokens for consistent processing across datasets.

\subsection{Layer Pruning Strategies}

\subsubsection{Activation-Based Pruning}

Activation-based pruning is a natural approach to identify redundant transformer layers due to the fundamental role activations play in neural network operations \cite{tushar2024activationbasedpruning}. Activations, measured as the output of neurons after applying nonlinear transformations, represent the input in a transformed feature space. Layers with specific activation patterns may contribute in various ways to the overall functionality of the network.

From a mathematical perspective, activations can be viewed as mappings from the input space to a feature space, where their magnitude and distribution signify the importance of a layer. Layers exhibiting consistently low or sparse activations are hypothesized to contribute minimally to overall feature transformation. Biologically, this aligns with the idea that neurons or brain regions with persistently low firing rates play a negligible role in processing, further motivating the use of activations as a basis for pruning \cite{harvey2013multiplexingstimulus}.

To quantify activations, we used three strategies to aggregate them into different signals or metrics: \textit{inhibition}, \textit{intensity}, and \textit{energy}. These metrics offer complementary insights into the importance of activations within a layer.

\textbf{Inhibition}: Inhibition measures the \textit{mean value of activations} in a layer:
\begin{equation}
    A_{\text{inhibition}} = \frac{1}{n \cdot d} \sum_{i=1}^n \sum_{j=1}^d A_{i,j}
\end{equation}
where \( A \in \mathbb{R}^{n \times d} \) is the activation matrix for \( n \) tokens and \( d \) hidden dimensions. Although this metric could also be termed \textit{polarity} (negative values indicate inhibition, positive values indicate excitation), we found it to be consistently negative across layers and datasets, leading to its designation as inhibition.

Although pruning inhibitory layers may occasionally be beneficial, it is often risky. Inhibitory layers may encode critical information, balance representations, or filter noise, ensuring efficient processing. Biologically, inhibitory neurons regulate excitatory activity, maintaining stability \cite{znamenskiy2024recurrentinhibition}. As a result, inhibition may not always be a reliable metric for pruning.

\textbf{Intensity}: Intensity measures the \textit{mean of absolute activations}, capturing the \( L_1 \)-norm:
\begin{equation}
    A_{\text{intensity}} = \frac{1}{n \cdot d} \sum_{i=1}^n \sum_{j=1}^d |A_{i,j}|
\end{equation}
Intensity reflects the magnitude of activation. Layers with low intensity often produce sparse activations, implying a limited influence on subsequent layers. Mathematically, low intensity reduces the transformation \( T(A) \) to bias terms:
\begin{equation}
    T(A) = W A + b \approx b
\end{equation}
Biologically, this aligns with the idea that neurons with weak signals contribute less to cognitive processing \cite{harvey2013multiplexingstimulus}. However, low-intensity layers can still encode selective and critical features, making pruning based solely on intensity occasionally misleading.

\textbf{Energy}: Energy measures the \textit{mean of squared values} of activations, capturing the \( L_2 \)-norm:
\begin{equation}
    A_{\text{energy}} = \frac{1}{n \cdot d} \sum_{i=1}^n \sum_{j=1}^d A_{i,j}^2
\end{equation}
Energy reflects the overall strength of the signal, with low energy suggesting that the layer has minimal influence on the network's computations. Energy is particularly useful because it magnifies larger activations while diminishing the impact of smaller ones. A low-energy layer produces outputs with minimal power, $\|A\|_2^2 = \text{Tr}(A^\top A)$. This suggests that low-power layers do not contribute substantially to overall information propagation. However, pruning based on energy may overlook layers that generate weak but highly structured signals essential for specific tasks, analogous to how certain brain regions exhibit low energy usage while maintaining critical functions.

Activation-based pruning methods offer compelling mathematical and biological rationales for identifying redundant layers. Inhibition, intensity, and energy provide diverse ways to quantify the contribution of activations. While these methods may succeed when activations are consistently low across various inputs, they may falter in cases where weak or sparse activations encode critical information.

\subsubsection{Mutual Information-Based Pruning}

Mutual information (MI) provides a rigorous framework for quantifying the dependence between variables, making it a natural candidate to evaluate the contribution of individual transformer layers \cite{isik2022informationtheoreticjustificationmodelpruning}. In the context of neural networks, MI captures how much information a layer's activations share with the target labels or adjacent layers. This enables principled pruning by identifying layers that contribute the least to task performance or exhibit high redundancy with neighboring layers.

From a mathematical perspective, MI measures the reduction in uncertainty about one variable given the knowledge of another. In transformers, activations at a given layer encode a representation of the input, and MI evaluates how much of this representation is task-relevant or novel compared to adjacent layers. Biologically, this approach aligns with the brain's reliance on efficient information transfer across neural circuits, where regions with low mutual information with their outputs or neighboring regions likely perform redundant or less critical computations \cite{xu2023mutualinformationvisual}.

To quantify the information contributed by each layer, we used two strategies to aggregate
them into different signals or metrics: \textit{Task-Relevance-MI} that computes MI between a layer and the target, and \textit{Flow-Relevance-MI} that computes MI flow between two consecutive layers. Each method offers unique insights into the contribution and redundancy of a layer based on shared information. Below, we provide mathematical definitions and analyze their implications for layer pruning.

\textbf{Task-Relevance-MI}: This strategy measures the dependency between a layer's activations and the target labels and informs whether a layer's output contributes task-critical information for prediction. Mathematically, the MI for layer \( l \) is defined as:
\begin{equation}
    \text{MI}_{\text{task}}(l) = I(A_l; y)
\end{equation}
where \( A_l \in \mathbb{R}^{n \times d} \) represents the activations of layer \( l \), and \( y \) denotes the target labels. \( I(A_l; y) \) is computed as:
\begin{equation}
    I(A_l; y) \approx \sum_{i=1}^n \log \frac{p(a_{l,i}, y_i)}{p(a_{l,i}) p(y_i)}
\end{equation}
where \( p(a_{l,i}, y_i) \) is the joint probability, and \(p(a_{l,i})\) and \(p(y_i)\) are marginal distributions.

Biologically, layers with high Task-Relevance-MI are analogous to brain regions specialized in processing task-relevant information (e.g. visual cortex for vision or somatosensory cortex for touch) \cite{rahman2019somatosentoryinteractions}, while layers with low Task-Relevance-MI are considered redundant and contribute minimally to specific tasks. Therefore, a low \( \text{MI}_{\text{task}}(l) \) suggests that the layer provides little task-relevant information and may be a candidate for pruning.

\textbf{Flow-Relevance-MI:} This strategy evaluates redundancy between adjacent layers, quantifying how much new information layer \( l+1 \) introduces relative to layer \( l \), and is defined as:
\begin{equation}
    \text{MI}_{\text{flow}}(l) = I(A_l; A_{l+1})
\end{equation}
where \( A_{l+1} \in \mathbb{R}^{n \times d} \) represents the activations of the subsequent layer. Since the activations in the intermediate layers are typically continuous, the computation of \(I(A_l; A_{l+1})\) can be estimated via the reduction in variance of \( A_l \) when conditioned on \( A_{l+1} \):
\begin{equation}
    I(A_l; A_{l+1}) \approx \text{Var}(A_l) - \text{Var}(A_l | A_{l+1})
\end{equation}
Biologically, layers with high Flow-Relevance-MI are analogous to brain regions that efficiently transfer information between interconnected areas, enabling hierarchical processing \cite{felleman1991distributedhierarchical}. In contrast, areas with low Flow-Relevance-MI are considered redundant and unnecessary (e.g., two nearly identical visual information processing circuits are not required and do not exist). Therefore, layers with low \( \text{MI}_{\text{flow}}(l) \) may be candidates for pruning.

In both methods, MI allows for targeted pruning decisions by identifying layers with low task relevance or high redundancy. However, these methods assume that low MI directly correlates with redundancy, which might overlook layers that encode intermediate features essential for downstream processing.

\subsubsection{Gradient-Based Pruning}

Gradient-based pruning uses the magnitude and structure of gradients to assess the contribution of individual transformer layers \cite{yang2023gradientbasedintraattentionpruningpretrained, molchanov2017pruningconvolutionalneuralnetworks}. Gradients, which represent the sensitivity of the loss function with respect to the model parameters, provide a direct measure of how much each layer contributes to reducing the loss. By analyzing gradient information, we can identify layers that exert minimal influence on the model's optimization dynamics and are thus potential candidates for pruning. 

From a mathematical perspective, gradients quantify the change in the model's output or loss in response to small perturbations in its parameters. Layers with consistently low gradient magnitudes indicate that their parameters are less significant for the optimization process and contribute minimally to performance improvement. Biologically, this aligns with the concept of synaptic plasticity in the brain, where connections with low or negligible weight updates over time are considered less critical for learning and can be pruned to improve efficiency \cite{magee2020synapticplasticity}.

To apply gradient-based pruning, we used two strategies to aggregate gradients into different signals or metrics: \textit{Gradient Magnitude}, which directly computes the magnitude the gradients, and \textit{Gradient Fisher Information}, which computes the variance of the derivative of the loss. Below, we provide mathematical definitions and analyze their implications for layer pruning.

\textbf{Gradient Magnitude:} This strategy computes the mean magnitude of the gradients for each layer, measuring the overall contribution of the layer to loss reduction. For a given layer \( l \), the gradient magnitude is defined as:
\begin{equation}
    G_{\text{magnitude}}(l) = \frac{1}{|\theta_l|} \sum_{p \in \theta_l} \left| \frac{\partial \mathcal{L}}{\partial \theta} \right|
\end{equation}
where \( \theta_l \) is the set of parameters in layer \( l \), \( \mathcal{L} \) is the loss function, and \( \frac{\partial \mathcal{L}}{\partial \theta} \) is the gradient of the loss with respect to parameter \( \theta \). Layers with low \( G_{\text{magnitude}}(l) \) suggest that their parameters contribute negligibly to reducing the loss, i.e., changing their parameters do not affect the loss much. So, those layers can be pruned.

\textbf{Fisher Information:} This strategy computed the expected change in the loss function when parameters are perturbed, providing a second-order measure of parameter importance. For a layer \( l \), the Fisher information is defined as:
\begin{equation}
    F(l) = \mathbb{E}_{(x, y) \sim \mathcal{D}} \left[ \left( \frac{\partial \mathcal{L}(x, y; \theta)}{\partial \theta_l} \right)^2 \right]
\end{equation}
where \( \mathcal{D} \) is the data distribution, \( \theta_l \) represents the parameters of layer \( l \), and \( \frac{\partial \mathcal{L}(x, y; \theta)}{\partial \theta_l} \) is the gradient of the loss with respect to \( \theta_l \). Fisher information highlights parameters or layers that are critical for maintaining the current loss minimum. Layers with low Fisher information imply that perturbing their parameters minimally affects the loss, making them redundant.

Gradient-based pruning strategies offer a direct measure of layer importance by evaluating their influence on loss optimization. Gradient magnitude provides an intuitive first-order measure, while Fisher information captures second-order effects, offering deeper insights into parameter significance. However, both methods rely on the assumption that low-gradient magnitudes or Fisher information correlate directly with redundancy. In practice, layers with low gradients might still play stabilizing roles, analogous to brain regions that act as modulatory hubs with minimal direct activity but essential indirect contributions.

\subsubsection{Weight-Based Pruning}

Weight-based pruning is another natural candidate to identify redundant transformer layers due to the foundational role weights play in defining layer transformations \cite{frankle2019lotterytickethypothesisfinding}. In neural networks, weights parameterize the linear mappings that transform inputs into feature representations, directly affecting the layer's contribution to the overall model. Layers with weak, sparse, or low-entropy weights are less likely to provide significant transformations, making them prime candidates for pruning without significantly impairing performance.

From a mathematical perspective, the weights \( W \in \mathbb{R}^{d_{\text{out}} \times d_{\text{in}}} \) define the transformations within a layer, where \( d_{\text{in}} \) and \( d_{\text{out}} \) represent the input and output dimensions. The properties of the weight matrix, such as its norm, sparsity, and entropy, provide key insights into the importance of a layer's contribution. Biologically, this aligns with the synaptic pruning mechanisms of the brain, where weak or redundant connections are systematically removed to optimize information processing \cite{paolicelli2011synapticpruning}.

To apply weight-based pruning, we used three strategies to aggregate the weights into different signals or metrics: \textit{norm}, \textit{sparsity}, and \textit{entropy}. These metrics provide various perspectives into the importance of weights within a layer. Below, each method is mathematically defined and analyzed in terms of its implications and effectiveness for layer pruning.

\textbf{Norm}: This strategy computes the \( L_2 \)-norm of the weight matrix in a layer, quantifying the overall magnitude of its parameters and capturing the strength of the transformation:
\begin{equation}
    \|W\|_2 = \sqrt{\sum_{i=1}^{d_{\text{out}}} \sum_{j=1}^{d_{\text{in}}} W_{i,j}^2}
\end{equation}
A low norm indicates that the layer's weights, \( W \), are close to zero, suggesting minimal contribution to the model's transformation. Mathematically, this implies that the layer's output is approximated primarily by its bias term:
\begin{equation}
\label{Weight-Norm-output-bias}
    T(A) = W A + b \approx b
\end{equation}
where \( T(A) \) is the transformed output of the layer, and \( A \) is the input activation from the previous layer. Layers with low-weight norms are considered redundant as they exert negligible influence on downstream computations. Consequently, such layers can be interpreted as weak connections that add little to the model's overall functionality, making them strong candidates for pruning.

\textbf{Sparsity}: This strategy measures the proportion of zero-valued elements in the weight matrix:
\begin{equation}
    S(W) = \frac{\sum_{i=1}^{d_{\text{out}}} \sum_{j=1}^{d_{\text{in}}} \mathbb{1}[W_{i,j} = 0]}{d_{\text{out}} \cdot d_{\text{in}}}
\end{equation}
where \( \mathbb{1}[\cdot] \) is the indicator function. A high sparsity value implies that most of the weights in the layer are zero. This implies that the output of the layer is approximated by the bias term as shown in \cref{Weight-Norm-output-bias}.

\textbf{Entropy}: This strategy measures the diversity in the weight distribution, reflecting the information content encoded by the weights. Mathematically,
\begin{equation}
    H(W) = -\sum_{i=1}^{d_{\text{out}}} \sum_{j=1}^{d_{\text{in}}} \frac{|W_{i,j}|}{\|W\|_1} \log \left( \frac{|W_{i,j}|}{\|W\|_1} + \epsilon \right)
\end{equation}
where \( \|W\|_1 = \sum_{i,j} |W_{i,j}| \) is the \( \ell_1 \)-norm of \( W \), and \( \epsilon \) is a small constant to avoid numerical instability. Layers with low entropy exhibit a highly concentrated weight distribution, dominated by a few large weights. Such layers may provide limited diversity in transformations, making them potential candidates for pruning. The weight entropy is analogous to the diversity of neural activation patterns in the brain. Circuits with concentrated activity are less efficient for generalizable tasks, whereas distributed activity allows for richer information processing.

Weight-based pruning strategies offer a mathematically sound and biologically inspired framework for identifying redundant layers. Although norm, sparsity, and entropy provide distinct insights into the significance of weights, pruning layers based on these metrics can fail if the underlying assumptions, i.e. low norm indicating low importance, high sparsity implying irrelevance, or low entropy signaling redundancy, do not accurately reflect the actual role of a layer in the network.

\subsubsection{Attention-Based Pruning}

Attention-based pruning uses the fundamental role of attention mechanisms in transformers to identify and remove redundant layers \cite{michel2019sixteenheadsreallybetter}. Attention weights indicate how tokens influence each other during the generation of contextual representations. Layers whose attention weights are uniformly distributed or consistently low are unlikely to capture meaningful token interactions, making them prime candidates for pruning with minimal impact on overall performance.

From a mathematical perspective, the attention weights \( \alpha \in \mathbb{R}^{n \times n \times h} \) quantify the influence of one token on another across \( h \) attention heads and \( n \) tokens. The properties of attention, such as their mean importance and entropy, provide insights into the relevance of a layer's attention mechanism. Biologically, this aligns with the concept of selective attention in neural circuits, where the brain prioritizes specific stimuli while suppressing others, ensuring efficient processing \cite{convento2018selectiveattention}. Similarly, layers with ineffective or redundant attention mechanisms in transformers can be pruned to optimize the model structure.

To quantify attention, we used two strategies to aggregate them into different signals or metrics: \textit{attention weight} and \textit{attention entropy}. These metrics offer complementary perspectives on the importance of attention mechanisms in a layer. Below, each method is mathematically defined and analyzed in terms of its implications and effectiveness for layer pruning.

\textbf{Attention Weight}: Attention weight measures the average magnitude of the attention scores across all tokens and heads within a layer:
\begin{equation}
    A_{\text{weight}} = \frac{1}{n^2 \cdot h} \sum_{i=1}^n \sum_{j=1}^n \sum_{k=1}^h \alpha_{i,j,k}
\end{equation}
where \( \alpha_{i,j,k} \) represents the attention score from token \( i \) to token \( j \) in head \( k \). A low attention weight suggests that the layer's attention mechanism assigns uniformly low importance across all tokens, indicating that the layer minimally influences the contextual representations. Mathematically, this implies that the layer contributes little to the model's ability to distinguish between relevant and irrelevant input tokens. Thus, such layers are strong candidates for pruning.

\textbf{Attention Entropy}: Attention entropy quantifies the diversity and concentration of attention scores, capturing the degree to which attention is focused or distributed:
\begin{equation}
\label{Attention-Entropy-equation}
    A_{\text{entropy}} = -\frac{1}{h} \sum_{k=1}^h \sum_{i=1}^n \sum_{j=1}^n \alpha_{i,j,k} \log \bigl(\alpha_{i,j,k} + \epsilon\bigr)
\end{equation}
where \(\epsilon\) is a small constant to prevent numerical instability. High entropy indicates that attention is evenly distributed across tokens, suggesting a lack of focus, while low entropy indicates concentrated attention on specific tokens. Biologically, this mirrors the brain's ability to focus selectively on critical stimuli while maintaining enough diversity to generalize across contexts. Thus, pruning high-entropy layers assumes that distributed attention contributes less to task-specific information flow.

Attention-based pruning methods provide a biologically plausible and mathematically rigorous approach to identifying redundant transformer layers. Attention weight and entropy offer complementary metrics for assessing layer relevance, with weight reflecting the overall magnitude of token interactions and entropy capturing their diversity. However, these methods may fail when uniformly low or distributed attention scores encode subtle but essential dependencies.

\subsubsection{Strategic Fusion}

Strategic fusion pruning methods combine individual strategies to make informed layer-pruning decisions. For a transformer model with $l$ layers, each layer is represented by a set of $m$ layer-specific signals. These signals form a feature matrix $\mathbf{X} \in \mathbb{R}^{l \times m}$, where each row $\mathbf{x}_l \in \mathbb{R}^m$ corresponds to the metrics of a specific layer $l$. We obtain 12 signals from 12 strategies for each layer, and therefore, $m=12$.

The importance of each layer is quantified by the target variable $\Delta \mathcal{A} \in \mathbb{R}^l$, where $\Delta \mathcal{A}_l$ represents the change in accuracy when a specific layer $l$ is pruned. Formally:
\begin{equation}
    \Delta \mathcal{A}_l = \mathcal{A}_{\text{orig}} - \mathcal{A}_l
\end{equation}

where $\mathcal{A}_{\text{orig}}$ is the accuracy of the original model and $\mathcal{A}_l$ is the accuracy after pruning layer $l$. A smaller $\Delta \mathcal{A}_l$ indicates that pruning the layer has a minimal impact on performance, making it a candidate for removal.

We introduced two independent fusion methods, linear regression and random forest. Both methods use the feature matrix $\mathbf{X} \in \mathbb{R}^{l \times m}$ and the corresponding accuracy change vector $\Delta \mathcal{A} \in \mathbb{R}^l$ to predict the pruning impact of each layer. These methods differ in their underlying assumptions and in the way they model relationships between strategies. During the iterative pruning process, both methods identify the layer $l^*$ with the lowest predicted impact:
\begin{equation}
\label{strategic-fusion-layer-selection}
    l^* = \arg\min_{l} \Delta \mathcal{A}_l
\end{equation}

This layer is then pruned, and the model is fine-tuned to adapt to the structural change. The process is repeated until a desired number of layers is pruned.

\textbf{Linear Regression-Based Pruning}: Linear regression assumes a linear relationship between the feature signals and the impacts, i.e. it measures impacts as a linear weighted combination of strategies. In this method, the target variable remains $\Delta \mathcal{A} \in \mathbb{R}^l$, while the feature space $\mathbf{X} \in \mathbb{R}^{l \times m}$ includes the same layer-specific signals. Linear regression model predicts pruning impact as:
\begin{equation}
\label{Linear-Regression-impact-prediction}
    \Delta \mathcal{A}_l \approx \mathbf{w}^\top \mathbf{x}_l + b
\end{equation}
where $\mathbf{w} \in \mathbb{R}^m$ are the learned weights indicating the significance of each metric, and $b$ is the bias.

\textbf{Random Forest-Based Pruning}: Random forest pruning provides a nonlinear way to quantify the importance of layers. Unlike linear regression, random forests capture complex relationships through an ensemble of decision trees. Each tree is trained on a random subset of the data, and the overall model aggregates predictions. In this method, the target variable remains $\Delta \mathcal{A} \in \mathbb{R}^l$, while the feature space $\mathbf{X} \in \mathbb{R}^{l \times m}$ includes the same layer-specific signals. The random forest model predicts pruning impact as:
\begin{equation}
\label{Random-Forest-impact-prediction}
    \Delta \mathcal{A}_l \approx \text{RF}(\mathbf{x}_l)
\end{equation}
where $\text{RF}(\mathbf{x}_l)$ represents the aggregated prediction from the ensemble. The model provides feature importance scores, which are analyzed to understand the relative contribution of individual strategies in the random forest fusion.

In both methods, the layer with the least predicted impact is pruned at each iteration, as guided by \cref{strategic-fusion-layer-selection}. The process continues until a target number of layers is pruned.

Linear regression and random forest are independent approaches for fusion-based pruning. Linear regression offers simplicity and mathematical clarity by assuming linear relationships, while random forest accounts for non-linear interactions, capturing more complex dependencies. Biologically, strategic fusion mirrors how different brain regions process different aspects of input at varying levels of complexity, collectively contributing to the final decision \cite{mesulam1998fromsensationtocognition, rahman2020auditoryandtactile}. This emphasizes the importance of integrating multiple strategies for robust and informed layer pruning.

\subsubsection{Random Pruning}
For comparison, we include a simple random pruning baseline in which each layer is selected uniformly at random for removal at each step, independent of any learned signals or metrics. This process is repeated until the desired number of layers has been pruned. To further confirm that an informed pruning sequence is critical for achieving optimal performance, we also repeated random pruning experiments on different dataset subsets, demonstrating that purely random selection consistently underperforms methods informed by layer-specific signals.

\subsection{Knowledge Distillation}
Layer pruning often leads to performance drop, e.g. accuracy, in the compressed model. To mitigate the accuracy drop after aggressive pruning, we used a knowledge distillation approach. We used the original (uncompressed) model as the teacher and the pruned model as the student. During training, the teacher produces soft probability distributions over classes for each input sample. The student model is then trained to mimic the teacher’s output distribution through a Kullback--Leibler divergence loss. Formally, let \(\mathbf{z}_t\) and \(\mathbf{z}_s\) be the logits of the teacher and student, respectively. We define the distillation loss \(\mathcal{L}_\text{KD}\) for a batch of size \(N\) as:

\begin{equation}
    \mathcal{L}_\text{KD} = \frac{1}{N} \sum_{i=1}^N 
    \mathrm{KL}\bigl(
        \sigma(\mathbf{z}_t^{(i)}/T) \;\big\|\; 
        \sigma(\mathbf{z}_s^{(i)}/T)
    \bigr)
\end{equation}

where \(\sigma(\cdot)\) denotes the softmax function and \( T \) denotes the temperature that determines the smoothness of the output distribution. We then combine \(\mathcal{L}_\text{KD}\) with the standard cross-entropy loss \(\mathcal{L}_\text{CE}\), computed using ground-truth labels, to form the overall training objective:

\begin{equation}
    \mathcal{L} = \alpha \,\mathcal{L}_\text{CE} + (1 - \alpha)\,\mathcal{L}_\text{KD}
\end{equation}

where \(\alpha\) controls the trade-off between adhering to the original labels and matching the soft output of the teacher. Empirically, this joint objective helps the student model absorb nuanced decision boundaries from the teacher, thereby recovering or even surpassing the accuracy lost through pruning. All distillation procedures follow the same training protocols used for fine-tuning, including learning rates, batch sizes, and optimizer settings, thus minimizing additional hyperparameter overhead. We used \( T = 2\) and \(\alpha=0.5\) in our experiments.

\subsection{Layer Pruning and Model Training}
Layer pruning involves removing specific transformer layers of a model to reduce its size and computational complexity while preserving performance. An identity wrapper is used to replace pruned layers in the model by acting as a placeholder. The wrapper is a module that simply passes the input through without performing any computations or transformations. This ensures that the overall architecture of the model remains intact and simplifies the implementation during the pruning process.

We used sequential pruning, which is an iterative approach that removes one layer at a time based on its importance level. The importance of a layer is determined by various signals generated by the layer. At each step, the importance of all layers is computed and the least important layer is removed. The pruned model is then fine-tuned with a small subset of training data and evaluated on the test data. The process is repeated until a desired number of layers is pruned. The same train and test subsets are used to fine-tune both the base model and the knowledge-distilled model. The fine-tuning process is identical for all models.

\begin{table*}[ht]
\caption{Maximum accuracy achieved by each method for each dataset. Bold values indicate the method that achieves the highest maximum accuracy for the corresponding dataset.}
\label{detailed-performance-table}
\vskip 0.15in
\begin{center}
\begin{small}
\resizebox{\textwidth}{!}{%
\begin{tabular}{l|c|c|c|c|c|c|c|c|c}
\toprule
\textbf{Strategy} & \textbf{newsgroup} & \textbf{dbpedia} & \textbf{arxiv} & \textbf{yahoo} & \textbf{patent} & \textbf{yelp} & \textbf{agnews} & \textbf{imdb} & \textbf{amazon} \\
\midrule
Baseline (uncompressed) & 0.289 & 0.977 & 0.367 & 0.262 & 0.320 & 0.406 & 0.832 & 0.705 & 0.723 \\
Random & 0.164 & 0.938 & 0.227 & 0.203 & 0.281 & 0.336 & 0.812 & 0.707 & 0.707 \\
Activation-Inhibition & 0.160 & 0.906 & 0.285 & 0.242 & 0.344 & 0.320 & 0.832 & 0.684 & 0.715 \\
Activation-Intensity & 0.117 & 0.875 & 0.266 & 0.184 & 0.246 & 0.379 & 0.773 & 0.695 & 0.719 \\
Activation-Energy & 0.117 & 0.891 & 0.254 & 0.188 & 0.254 & 0.371 & 0.766 & 0.684 & 0.727 \\
Task-Relevance-MI & 0.191 & 0.938 & 0.312 & 0.242 & 0.305 & 0.367 & 0.816 & 0.699 & \textbf{0.730} \\
Flow-Relevance-MI & 0.117 & 0.941 & 0.312 & 0.184 & 0.336 & 0.344 & 0.836 & 0.688 & 0.699 \\
Gradient-Magnitudes & 0.117 & 0.930 & 0.207 & 0.184 & 0.254 & 0.344 & 0.812 & 0.691 & 0.695 \\
Gradient-Fisher & 0.109 & 0.906 & 0.320 & 0.227 & \textbf{0.352} & 0.379 & 0.812 & 0.715 & \textbf{0.730} \\
Weight-Norm & 0.098 & 0.918 & 0.238 & 0.191 & 0.219 & 0.355 & 0.770 & 0.688 & 0.707 \\
Weight-Sparsity & 0.129 & 0.922 & 0.227 & 0.215 & 0.289 & 0.328 & 0.832 & 0.699 & 0.715 \\
Weight-Entropy & 0.215 & 0.934 & 0.324 & 0.234 & 0.336 & 0.371 & 0.828 & 0.695 & 0.711 \\
Attention-Weight & 0.188 & 0.93 & 0.289 & 0.215 & 0.328 & 0.363 & 0.832 & 0.703 & \textbf{0.730} \\
Attention-Entropy & 0.164 & 0.879 & 0.258 & 0.219 & 0.324 & 0.348 & 0.805 & 0.695 & 0.734 \\
Linear-Regression & \textbf{0.508} & \textbf{0.945} & 0.305 & 0.230 & 0.344 & \textbf{0.395} & 0.836 & \textbf{0.711} & \textbf{0.730} \\
Random-Forest & \textbf{0.508} & \textbf{0.945} & \textbf{0.328} & \textbf{0.250} & 0.336 & 0.383 & \textbf{0.840} & \textbf{0.711} & \textbf{0.730} \\
\bottomrule
\end{tabular}%
}
\end{small}
\end{center}
\vskip -0.1in
\end{table*}

\begin{figure}[!t]
\vskip 0.2in
\begin{center}
\centerline{\includegraphics[width=\columnwidth]{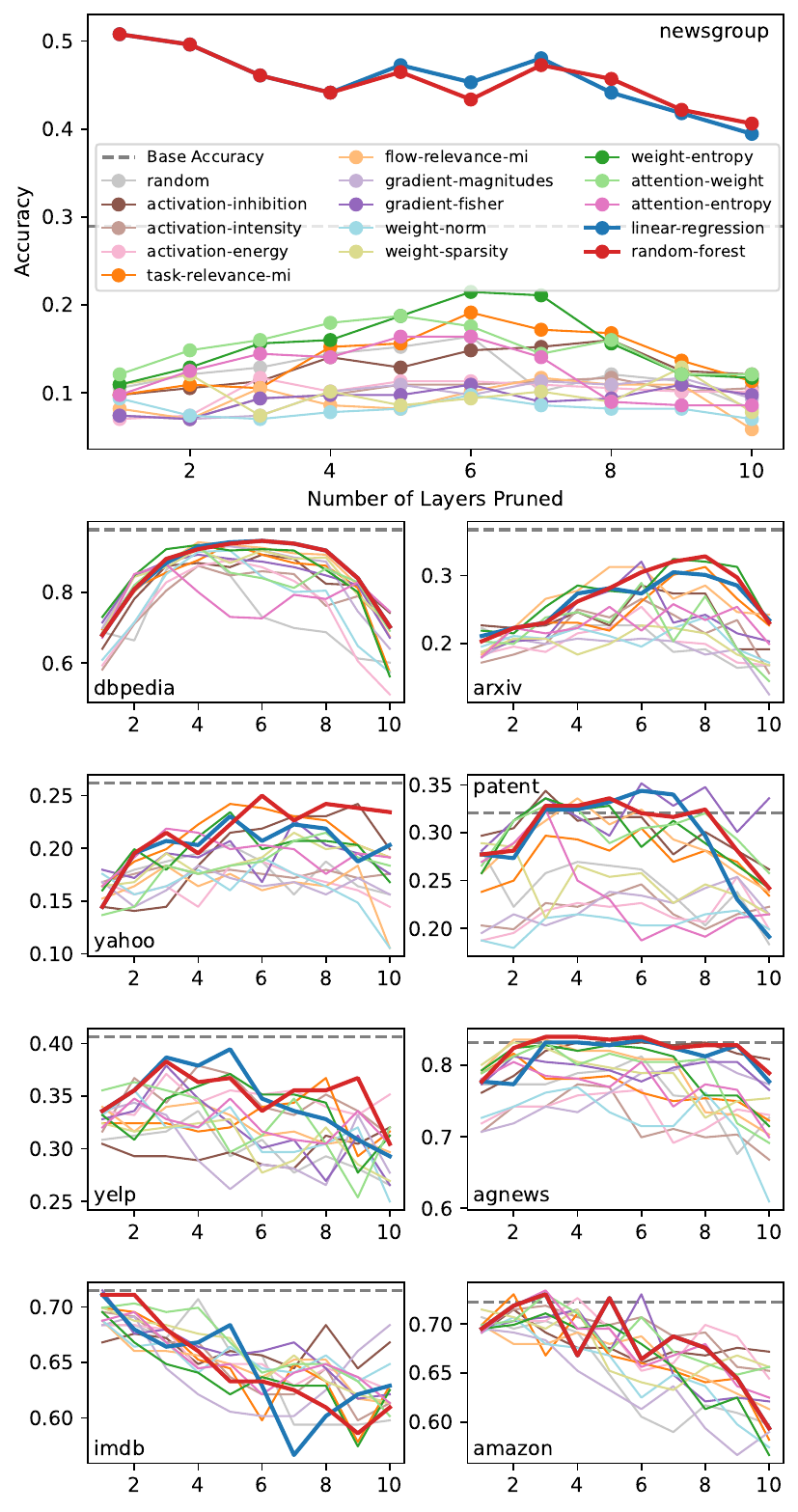}}
\caption{Comparison of accuracies across nine datasets as the increasing number of transformer layers being pruned. Each colored line corresponds to a distinct pruning strategy, with the dashed line indicating the unpruned baseline accuracy. The plots highlight how different pruning criteria affect model performance at varying compression levels.}
\label{accuracy-trends-vs-layers-pruned}
\end{center}
\vskip -0.2in
\end{figure}

\begin{figure}[ht]
\vskip 0.2in
\begin{center}
\centerline{\includegraphics[width=\columnwidth]{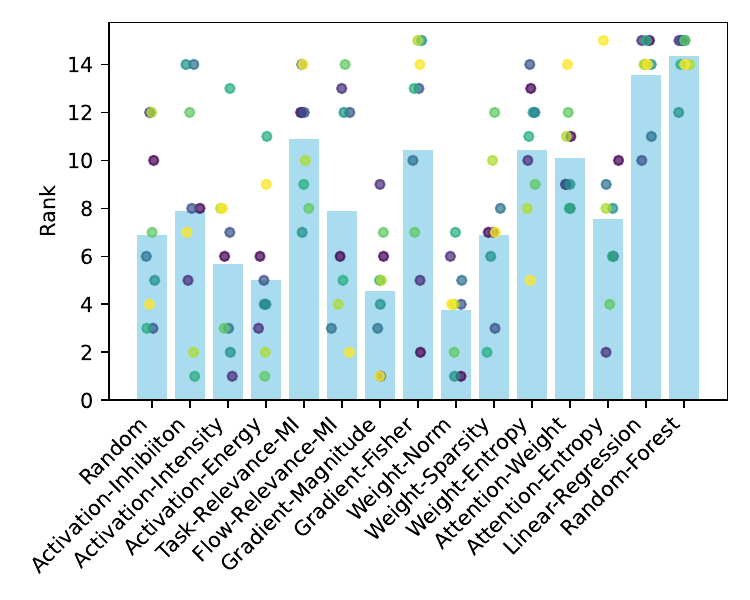}}
\caption{Ranking of the strategies for various datasets. Bars represent the mean rank of each method, while dots indicate the rank for individual datasets. Ranks are computed by sorting strategies for each dataset based on their maximum accuracy, with the highest accuracy assigned rank 15 (because of a total of 15 strategies including random), the second highest rank 14, and so on.}
\label{method-ranks}
\end{center}
\vskip -0.2in
\end{figure}

\begin{figure}[ht]
\vskip 0.2in
\begin{center}
\centerline{\includegraphics[width=\columnwidth]{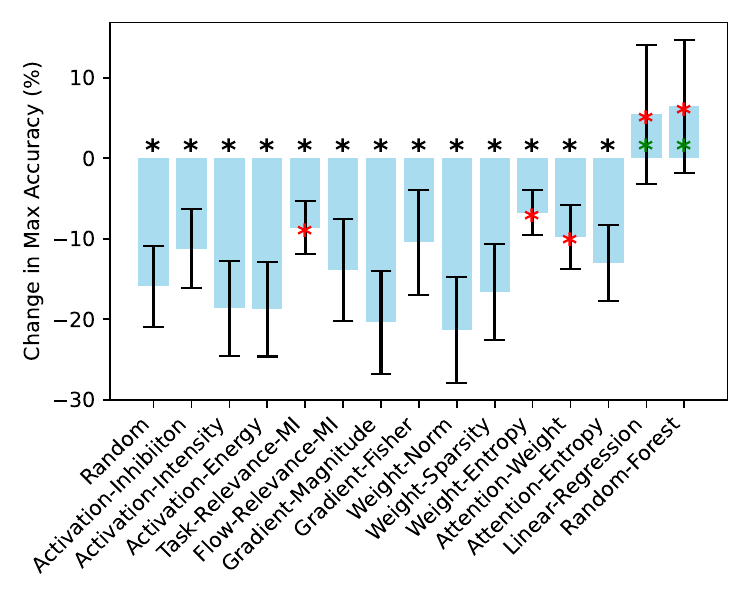}}
\caption{Percentage change in maximum accuracy compared to the baseline for each strategy. Black asterisks indicate that the means are significantly less than zero. Red asterisks indicate that the means are significantly higher than the mean of random method. Green asterisks indicate that the means are not significantly different from zero. All tests are based on Wilcoxon signed-rank test, \( p < 0.05 \).}
\label{max-accuracy-to-base-accuracy-comparison}
\end{center}
\vskip -0.2in
\end{figure}

% \FloatBarrier

\begin{figure}[ht]
\vskip 0.2in
\begin{center}
\centerline{\includegraphics[width=\columnwidth]{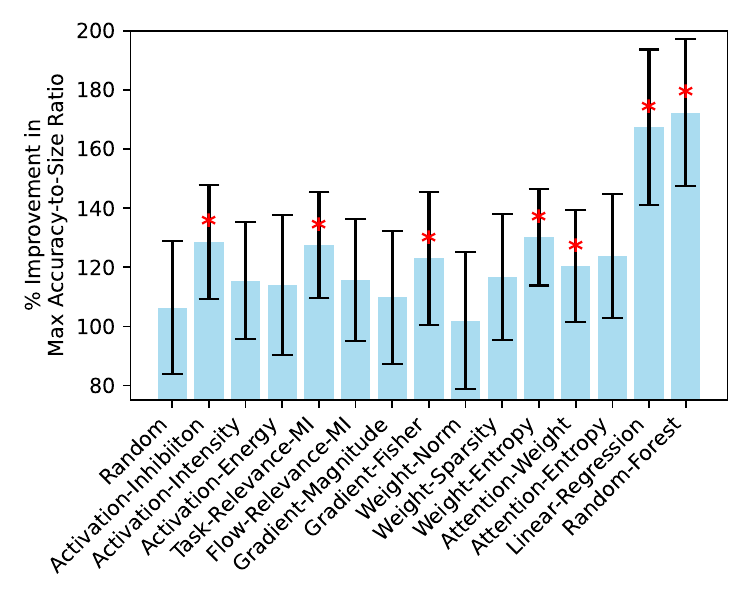}}
\caption{Maximum accuracy-to-size ratio for each method, averaged across all datasets. Error bars represent the standard error of the mean. Red asterisks indicate strategies in which the ratio is statistically significantly different from the random strategy (Wilcoxon signed-rank test, \( p < 0.05 \)).}
\label{max-accuracy-to-size-ratio-vs-methods}
\end{center}
\vskip -0.2in
\end{figure}

\begin{figure}[ht]
\vskip 0.2in
\begin{center}
\centerline{\includegraphics[width=\columnwidth]{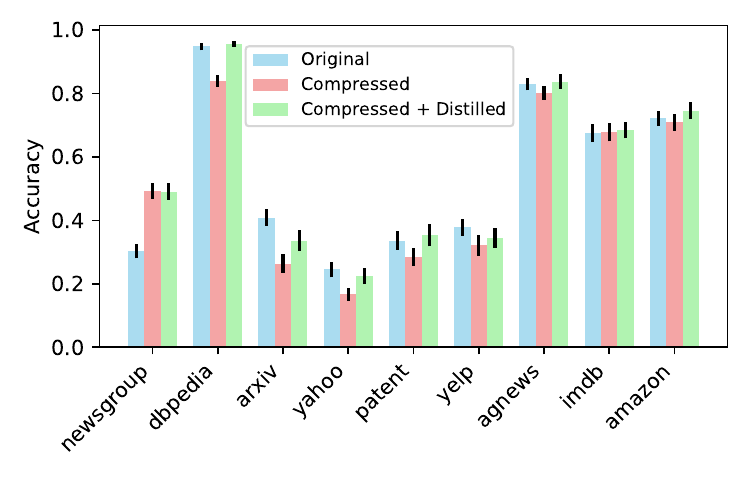}}
\caption{Accuracy comparison between the original model, compressed model (random forest strategic fusion), and compressed model with knowledge distillation. Distillation surpasses original accuracy for six datasets, and mitigates accuracy drops in the remaining three.}
\label{accuracy-with-knowledge-distillation}
\end{center}
\vskip -0.2in
\end{figure}

\begin{figure}[ht]
\vskip 0.2in
\begin{center}
\centerline{\includegraphics[width=\columnwidth]{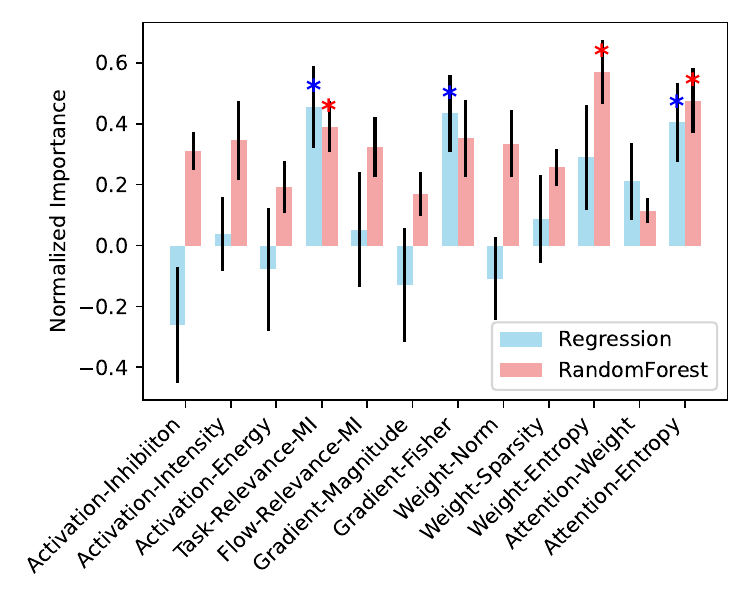}}
\caption{Normalized importance of each method, averaged across all datasets. Errorbar indicates standard error of mean. Blue asterisk indicates that the weight assigned to the strategy/feature by linear regression is significantly higher than zero. Red asterisk indicates that the importance of the strategy in random forest is significantly higher than the median importance across all strategies. All tests are based on Wilcoxon signed-rank test, \(p < 0.05\).}
\label{normalized-importance-vs-methods}
\end{center}
\vskip -0.2in
\end{figure}

\begin{figure}[ht]
\vskip 0.2in
\begin{center}
\centerline{\includegraphics[width=\columnwidth]{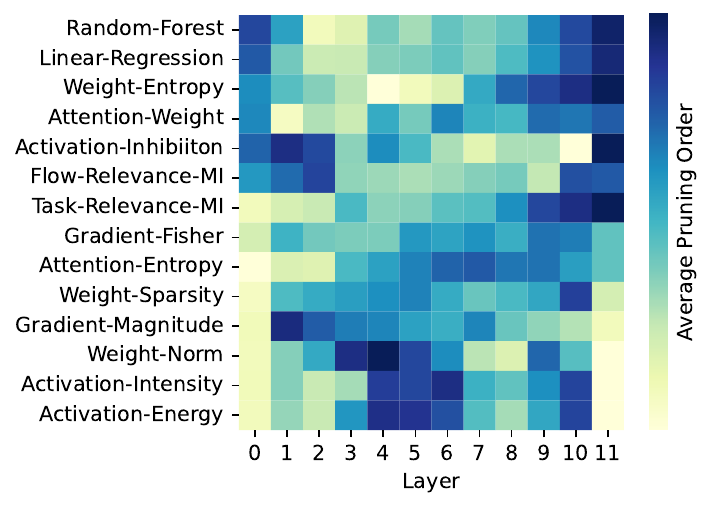}}
\caption{Heatmap of average order in which each layer was pruned for various pruning strategies. Darker colors indicate layers pruned later in the sequence, while lighter colors represent layers pruned earlier.}
\label{layers-ranks-by-methods}
\end{center}
\vskip -0.2in
\end{figure}

\begin{figure}[ht]
\vskip 0.2in
\begin{center}
\centerline{\includegraphics[width=\columnwidth]{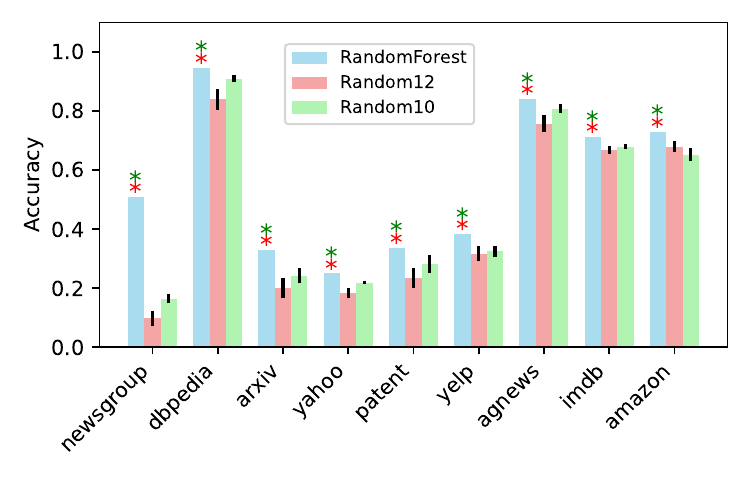}}
\caption{Layer randomization tests. The blue bars represent the maximum accuracy obtained using the RandomForest strategic fusion. The red bars represent the mean of maximum accuracies obtained by randomizing 12 layers (Random12), while the green bars represent the mean of maximum accuracies obtained by randomizing 10 layers, excluding the first and last layers (Random10). Red asterisks indicate that the maximum accuracy of RandomForest is significantly higher than the mean of Random12, while green asterisks indicate that the maximum accuracy of RandomForest is significantly higher than the mean of Random10 for the same dataset. All tests are based on the Wilcoxon signed-rank test, $p < 0.05$.}
\label{layer-randomization-tests}
\end{center}
\vskip -0.2in
\end{figure}

\section{Results}

We investigated 14 distinct layer pruning strategies, organized into six categories, activation, mutual information, gradient, weight, attention, and strategic fusion, alongside a random pruning baseline. We evaluated each strategy using nine datasets. This section presents how each strategy reduces model size while retaining (or even surpassing) performance, and highlights the advantages of integrating multiple layer-specific signals in a unified pruning framework.

\subsection{Strategic Fusion Optimizes Model Compression}

\cref{accuracy-trends-vs-layers-pruned} illustrates the accuracy trends as the model is sequentially compressed by pruning layers, one at a time, across nine datasets. For the newsgroup dataset, the trends derived from strategic fusion methods (Random Forest and Linear Regression) are notably distinct from the others. Other well-performing methods include Weight-Entropy, Task-Relevance-MI, and in some datasets, Gradient-Fisher. Although the performance of different strategies varies across the datasets, the general trends of high-performing strategies remain consistent.

The trends in \cref{accuracy-trends-vs-layers-pruned} provide a comprehensive view of how each strategy performs as the layers are pruned sequentially across the datasets. However, the ultimate goal is to identify a model, with a specific number of pruned layers, that performs optimally. To this end, we extracted the maximum accuracy achieved by each strategy for each dataset, as summarized in \cref{detailed-performance-table}. The results reveal that the strategic fusion with the random forest performs best for seven out of nine datasets, while the strategic fusion with linear regression leads for five datasets. Other notable methods, such as Task-Relevance-MI, Gradient-Fisher, and Attention-Weight, also perform well in specific datasets. Interestingly, the highest accuracies achieved by these individual strategies are also obtained through the strategic fusion methods, with the exception of Gradient-Fisher for one dataset. This indicates that strategic fusion-based layer pruning consistently outperforms individual strategy-based pruning, highlighting the importance of considering interactions between layer-specific metrics for informed pruning decisions. Moreover, the superior performance of the random forest-based fusion suggests that nonlinear interactions between these metrics are more prevalent and impactful than linear ones.

So far, we have focused on maximum accuracy as a measure of effectiveness for the best-performing model on each dataset. However, relying solely on maximum accuracy overlooks three critical factors. First, strategies that yield accuracies close to the maximum, such as the second or third best, can also be effective and merit consideration, as small differences in accuracy may not translate into meaningful performance differences. Second, this approach ignores the accuracy drops from the baseline model, which are crucial for assessing the feasibility of pruning. Ideally, pruning should result in a negligible or no accuracy drop; substantial drops could make pruning unsuitable. Third, some strategies might achieve slightly lower accuracy but with more layers pruned, resulting in smaller models, a trade-off that is essential for many resource-constrained applications. Addressing these factors allows for a more nuanced evaluation of the strategies and their practical effectiveness.

We addressed the first factor by computing the rank of each strategy for each dataset. For a given dataset, the strategies are ranked based on maximum accuracy and the rank (index + 1) is assigned to each strategy. These ranks, displayed in \cref{method-ranks}, reveal that the random forest consistently achieves high ranks in all datasets, followed by linear regression. Both methods are clearly distinguishable from other individual strategies. Although certain datasets rank individual strategies higher, such as Gradient-Fisher, Task-Relevance-MI, Weight-Entropy, and Attention-Weight, most datasets rank these independent strategies lower overall.

To address concerns about decline in accuracy, we calculated the change in maximum accuracy relative to baseline accuracy, as shown in \cref{max-accuracy-to-base-accuracy-comparison}. Independent strategies show consistent accuracy drops, with their accuracies statistically lower than zero (Wilcoxon signed-rank test, \( p < 0.05 \)). In contrast, the changes in accuracy in strategic fusion approaches, both linear regression and random forest, are not significantly different from zero (\( p < 0.05 \)), indicating that strategic fusion maintains baseline accuracy. Among individual strategies, Task-Relevance-MI, Weight-Entropy, and Attention-Weight showed smaller accuracy drops than the random method (\( p < 0.05 \)), demonstrating their relative effectiveness.

To address both accuracy and model size considerations, we computed the percentage improvement in the maximum accuracy-to-size ratio by comparing the ratio of the maximum accuracy to the size of the compressed model with the corresponding ratio for the base model. This improvement reflects the relative percentage increase in the accuracy-to-size ratio of the compressed model compared to the original uncompressed model, highlighting the efficiency gained through pruning. The size of the model is determined by the number of parameters it contains. The original uncompressed model has 109,489,930 parameters, and pruning one layer reduces the number of parameters by 7,087,872. These parameters are converted to gigabytes assuming each weight uses 32-bit precision. This sequential reduction in parameters during pruning is factored into the size calculations, providing a basis for evaluating the trade-off between accuracy and size.

This metric, shown in \cref{max-accuracy-to-size-ratio-vs-methods}, indicates that the mean value of this metric is significantly higher for the two strategic fusion approaches (Linear-Regression and Random-Forest) and for five individual strategies (Activation-Inhibition, Task-Relevance-MI, Gradient-Fisher, Weight-Entropy, and Attention-Weight) compared to the random method (\( p < 0.05 \)). However, as the figure shows, the maximum accuracy-to-size ratio achieved by strategic fusion approaches is distinctly higher and clearly outperforms the independent strategies, highlighting the superior performance of strategic fusion in balancing accuracy and compression.

\subsection{Knowledge Distillation Mitigates Accuracy Drops}
We have established that strategic fusion-based layer pruning, specifically with Random-Forest, is an effective compression technique. However, as with other compression methods, this approach often reduces the accuracy compared to the original model, with some rare exceptions. One way to mitigate this accuracy drop is through knowledge distillation, where the compressed model (student) is trained to mimic the predictions of the original model (teacher). During this process, the teacher model provides "soft labels" (probabilistic outputs) as guidance, which helps the student model learn finer-grained information about the data distribution beyond hard labels.

To evaluate this, we created a student model based on the compressed model that achieved the highest accuracy using the random forest-based layer pruning strategy. We then trained the student model using the teacher model and compared the accuracies of the original model, the compressed model, and the compressed model with distillation. These results are shown in \cref{accuracy-with-knowledge-distillation}. In most datasets, the compressed model exhibited a drop in accuracy compared to the original model, except for newsgroup and imdb, where the compressed model performed with higher accuracies. After applying knowledge distillation, the accuracy of the compressed model improved in most cases and, in some instances, even surpassed the original model's accuracy. Using accuracies after distillation, the accuracy-to-size ratio increased by a factor of 18.84 on average (mean: 18.84, std: 6.28, min: 10.29, max: 29.29).

Our results align with the findings by \cite{muralidharan2024compactlanguagemodelspruning}, who showed that pruning combined with selective retraining achieves state-of-the-art compression with minimal performance degradation. However, the improvement is not evident for datasets in which the compressed model already significantly outperformed the original model. This suggests that the teacher model may have limitations in effectively transferring knowledge to the student model in these scenarios. Therefore, while random forest-based compression followed by knowledge distillation effectively mitigates accuracy drops, distillation may not be necessary when the compressed model already achieves higher accuracy.

\subsection{Why Does Strategic Fusion Outperform Individual Strategies?}

In this section, we address why strategic fusion performs better than individual strategies. Strategic fusion combines multiple strategies to capture the significance of each layer in a more comprehensive way. Linear regression aggregates individual strategies using a linear weighted combination, while random forest captures the nonlinear interactions between the strategies. This allows fusion models to incorporate multiple layer-specific signals and ensures that no single strategy or signal dominates the pruning decision. The aggregated metric essentially uses the strengths of individual strategies, compensating for their potential weaknesses, and leads to better-informed pruning decisions.

We analyze different strategies from different perspectives. First, we measure how much each underlying strategy contributes to the fusion models. Specifically, we retrieve the learned weights (linear regression) or feature importance (random forest) assigned to each strategy, rescale them in \([-1, +1]\), and average them across datasets as shown in \cref{normalized-importance-vs-methods}. We observe that Task-Relevance-MI and Gradient-Fisher are dominant in the linear regression fusion, while Task-Relevance-MI and Weight-Entropy contribute more to the random forest fusion. Intriguingly, Attention-Entropy plays a significant role in both fusion strategies, but did not appear among top performers in other analyses (\cref{detailed-performance-table,method-ranks,max-accuracy-to-size-ratio-vs-methods}). This finding is a good example of why single-metric pruning, such as strategies with attention entropy or activation energy, can fail.

Attention entropy (defined in \cref{Attention-Entropy-equation}) measures how broadly or narrowly attention is spread. Ignoring the small constant \(\epsilon\) for simplicity, we have:
\begin{equation}
    A_{\text{entropy}} 
    = 
    -\frac{1}{h} 
    \sum_{k=1}^h \sum_{i=1}^n \sum_{j=1}^n 
    \alpha_{i,j,k} 
    \log \bigl(\alpha_{i,j,k}\bigr)
\end{equation}
where $\alpha_{i,j,k}$ is the attention score from token $i$ to token $j$ for head $k$. For each specific \(i\) and \(k\):
\begin{equation}
    A_{\text{entropy}}^{i,k} 
    = 
    -\sum_{j=1}^n 
    \alpha_{i,j,k} 
    \log \bigl(\alpha_{i,j,k}\bigr)
\end{equation}

High entropy arises when attention is evenly distributed (\(\alpha_{i,j,k} = 1/n\)), giving \(A_{\text{entropy}}^{i,k}=\log n\). Low entropy arises when attention is entirely focused on a single token (\(\alpha_{i,j_*,k} = 1 \text{ and } \alpha_{i,j,k} = 0 \text{ for } j \neq j_*\)), giving \(A_{\text{entropy}}^{i,k}=0\). Accordingly, the attention output:
\begin{equation}
    z_{i,k}=\sum_{j=1}^n \alpha_{i,j,k} \, V_{j,k}
\end{equation}

where $V_{j,k}$ is the value vector for token $j$ in head $k$, becomes:
\begin{equation}
\label{attention-output-high-entropy}
    z_{i,k}=\frac{1}{n}\sum_{j=1}^n V_{j,k} \text{ (for high entropy) }
\end{equation}
and
\begin{equation}
\label{attention-output-low-entropy}
    z_{i,k}=V_{j_*,k} \text{ (for low entropy)}
\end{equation}
\cref{attention-output-high-entropy} implies that, with high entropy, no single token receives a higher score than others, so the heads capture broader context and miss key details. Conversely, \cref{attention-output-low-entropy} implies that, with low entropy, the heads are entirely focus on token $j_*$ and neglect the context from other tokens. Both extremes are suboptimal, as an effective balance between context and focus is crucial. A single-signal pruning strategy, based solely on attention entropy, tends to favor one of these extremes. Consequently, such a strategy is unlikely to perform well, whether it prioritizes pruning low-entropy or high-entropy layers first. In contrast, fusion-based strategies adaptively achieve this balance by considering interactions with other metrics.

Second, we investigate the similarities and differences between the strategies, and get an insight why one might perform better than the other. We use the sequence of layers pruned by each strategy for this investigation. We computed the importance ranks of the layer as the order of sequence of pruning, e.g., if a layer is pruned first by a strategy, the rank is 1, while if a layer is pruned second, the rank is 2, and so on. \cref{layers-ranks-by-methods} demonstrates the average ranking of layer importance for various strategies. High-performing strategies, such as Random-Forest and Linear-Regression, rank the edge layers (beginning and ending layers) higher. Top-performing individual strategies, such as Weight-Entropy, Attention-Weight, and Activation-Inhibition, also rank the edge layers higher. Other top-performing individual strategies, such as Task-Relevance-MI and Gradient-Fisher, rank the ending layers higher. In contrast, poorly performing strategies, such as Weight-Norm, and Activation-Energy, consistently assign lower ranks to the both edge layers and prune them early. The starting layers extract low-level features from the input, and the ending layers translate these into task-specific outputs. Therefore, any strategy that prunes the edge layers early can disrupt critical functionality and lead to accuracy degradation.

Many metrics may exhibit consistently low or high values in the edge layers. Single-metric strategies rely on predefined patterns (e.g., pruning layers with low metric values first, or high metric values first) to determine the pruning order. However, these predefined patterns do to account for the contextual significance of the layers, such as their functional roles in the model. For example, the output layer is specialized for task-specific processing. For a given task, only a few neurons may exhibit high activity, while many outputs remain close to zero. This can result in low values for metrics such as activation energy, activation intensity, or weight norm. However, these low metric values should not imply that the output layer is less important, as it plays a crucial role in translating the learned features into task-specific predictions. Fusion methods can avoid this issue by combining multiple metrics, allowing the model to account for the functional importance of layers rather than relying on any predefined rules or on individual metric values.

Third, the fusion model is more analogous to brain circuits than a single strategy. Fusion strategy mirrors how the brain integrates diverse signals to make decisions \cite{mazurek2003neuralintegrators}. The human brain does not rely on a single cue, but instead combines information from multiple sources, such as sensory inputs, contextual relevance, and past experiences, to prioritize and allocate resources effectively. Similarly, fusion strategies aggregate multiple individual strategies, each reflecting a distinct aspect of layer importance. This fusion enables a comprehensive assessment of which layers to prune first and next. While this analogy is not data-driven and is based on what is known about brain functions, it does not provide definitive conclusions about pruning strategies. Instead, it serves as an inspiration to guide further research in exploring biologically inspired approaches to model optimization.

\subsection{Informed Sequencing of Layer Pruning is Essential for Optimal Performance}
In this research, we proposed that strategic fusion with a random forest represents an optimal strategy, where layers are pruned sequentially based on optimal information. However, a key question remains: does the informed sequence matter, or can layers be pruned in any order? To address this, we conducted two experiments. Both experiments followed the same approach as the random forest strategy, with one key difference. In the first experiment, the layers are randomly selected in each iteration. In the second experiment, the layers are randomly selected, excluding the first and last layers. Each experiment is repeated ten times and the mean of the maximum accuracies is computed, as shown in \cref{layer-randomization-tests}. The results show that the accuracy achieved by the Random-Forest strategic fusion is statistically significantly higher than that of either experiment. This suggests that an informed sequencing of layer pruning is essential for optimal performance. Additionally, while the two edge layers are critical, well-informed sequencing of the mid-layers also plays a vital role.

\section{Conclusions}
In this study, we explored and evaluated 14 layer pruning strategies using nine text datasets. Twelve single-metric strategies use signals obtained from layer activations, gradients, mutual information, weights, and attention. Two strategic fusion strategies, grounded in linear regression and random forests, combine individual strategies to automatically learn the pruning sequence in a data-driven framework. For each strategy, we provided the mathematical and biological intuition behind their choice and demonstrated theoretical and practical perspectives on their selection. We employed sequential pruning to iteratively remove the least critical layer identified by each strategy. Finally, we adopted knowledge distillation to mitigate any performance drop from the pruning.

Our key contribution is the strategic fusion framework that reconciles multiple pruning signals into a unified decision criterion. Specifically, our random forest-based approach outperformed individual strategies in both accuracy and accuracy-to-size ratio metrics. Notably, knowledge distillation exceeded the original accuracy for most datasets and mitigated the accuracy drop for other datasets during the pruning process. These findings demonstrate a practical path toward aggressive yet accurate transformer model compression.

We demonstrated the limitations of single-metric-based strategies, such as attention entropy and activation energy. Our insights were supported by the mathematical and functional foundations of the metrics, along with an analysis of the sequence of layers pruned. We highlighted how fusion-based strategies address the limitations of single-metric approaches. To further support our approach, we drew analogies with biological circuits, and illustrated how these systems inform and align with our various strategies. These mathematical and biological analogies provide both theoretical and practical perspectives on the selection of strategies. They also present a compelling case for exploring biologically inspired approaches to model optimization in future research.

Despite promising results and prospects, our framework has a few limitations. First, each pruning step requires fine-tuning to adapt the remaining model parameters, which can be an expensive process for large-scale datasets \cite{wu2024efficiencyscalingaisustainably}. Second, our approach focuses on pruning entire layers without exploring more granular compression (e.g., token or head pruning) \cite{kim2022learnedtokenpruningtransformers, michel2019sixteenheadsreallybetter}. Finally, although random forest demonstrates the best performance in most cases, it may not always be the optimal choice for new datasets (\cref{detailed-performance-table}). A more robust approach would be to complement random forest fusion with linear regression fusion and, whenever possible, with top-performing individual strategies, such as gradient Fisher information or task-relevant mutual information. These limitations, however, present opportunities for further research in the field.

Overall, our research highlights the transformative potential of strategic fusion that combines multiple pruning signals within a data-driven framework. Such a framework can effectively maintain or even exceed the original model accuracy while significantly improving the accuracy-to-size ratio. By systematically balancing accuracy and model size, this work establishes a foundation for deploying transformer models more effectively in resource-constrained, real-world applications.

\section*{Impact Statement}

This paper presents work whose goal is to advance the field of Machine Learning. There are many potential societal consequences  of our work, none which we feel must be specifically highlighted here.

% In the unusual situation where you want a paper to appear in the
% references without citing it in the main text, use \nocite
% \nocite{langley00}

\bibliographystyle{icml2025}
\bibliography{icml2025}

%%%%%%%%%%%%%%%%%%%%%%%%%%%%%%%%%%%%%%%%%%%%%%%%%%%%%%%%%%%%%%%%%%%%%%%%%%%%%%%
%%%%%%%%%%%%%%%%%%%%%%%%%%%%%%%%%%%%%%%%%%%%%%%%%%%%%%%%%%%%%%%%%%%%%%%%%%%%%%%
% APPENDIX
%%%%%%%%%%%%%%%%%%%%%%%%%%%%%%%%%%%%%%%%%%%%%%%%%%%%%%%%%%%%%%%%%%%%%%%%%%%%%%%
%%%%%%%%%%%%%%%%%%%%%%%%%%%%%%%%%%%%%%%%%%%%%%%%%%%%%%%%%%%%%%%%%%%%%%%%%%%%%%%
\newpage
\appendix
\onecolumn
% \section{You \emph{can} have an appendix here.}

%%%%%%%%%%%%%%%%%%%%%%%%%%%%%%%%%%%%%%%%%%%%%%%%%%%%%%%%%%%%%%%%%%%%%%%%%%%%%%%
%%%%%%%%%%%%%%%%%%%%%%%%%%%%%%%%%%%%%%%%%%%%%%%%%%%%%%%%%%%%%%%%%%%%%%%%%%%%%%%

\end{document}